\documentclass[conference]{IEEEtran}
\IEEEoverridecommandlockouts
\usepackage{cite,hyperref,multirow}
\usepackage{amsmath,amssymb,amsfonts}
\usepackage{algpseudocode}
\usepackage{algorithm}
\usepackage{tikz}
\usepackage{graphicx}
\usepackage{textcomp}
\usepackage{xcolor}
\def\BibTeX{{\rm B\kern-.05em{\sc i\kern-.025em b}\kern-.08em
    T\kern-.1667em\lower.7ex\hbox{E}\kern-.125emX}}
\begin{document}

\title{Tokenization Matters: Improving Zero-Shot NER for Indic Languages}
\author{\IEEEauthorblockN{1\textsuperscript{st} Priyaranjan Pattnayak}
\IEEEauthorblockA{\textit{University of Washington} \\
Seattle \\
 ppattnay@uw.edu
}
\and

\IEEEauthorblockN{2\textsuperscript{nd} Hitesh Patel}
\IEEEauthorblockA{\textit{New York University} \\
New York \\
hitesh.patel945@gmail.com
}
\and

\IEEEauthorblockN{3\textsuperscript{rd} Amit Agarwal}
\IEEEauthorblockA{\textit{Liverpool John Moores University} \\
Liverpool \\
amit.pinaki@gmail.com 
}
}

\maketitle

\begin{abstract}
Tokenization is a critical component of Natural Language Processing (NLP), especially for low-resource languages, where subword segmentation influences vocabulary structure and downstream task accuracy. Although Byte-Pair Encoding (BPE) is a standard tokenization method in multilingual language models, its suitability for Named Entity Recognition (NER) in low-resource Indic languages remains underexplored due to its limitations in handling morphological complexity. In this work, we systematically compare BPE, SentencePiece, and Character-Level tokenization strategies using IndicBERT for NER tasks in low-resource Indic languages like Assamese, Bengali, Marathi, and Odia, as well as extremely low-resource Indic languages like Santali, Manipuri, and Sindhi. We assess both intrinsic linguistic properties—tokenization efficiency, out-of-vocabulary (OOV) rates, and morphological preservation—as well as extrinsic downstream performance, including fine-tuning and zero-shot cross-lingual transfer.

Our experiments show that SentencePiece is a consistently better performing approach than BPE for NER in low-resource Indic Languages, particularly in zero-shot cross-lingual settings, as it better preserves entity consistency. While BPE provides the most compact tokenization form, it is not capable of generalization because it misclassifies or even fails to recognize entity labels when tested on unseen languages. In contrast, SentencePiece constitutes a better linguistic structural preservation model, benefiting extremely low-resource and morpholically rich Indic languages, such as Santali and Manipuri, for superior entity recognition, as well as high generalization across scripts, such as Sindhi, written in Arabic. The results point to SentencePiece as the more effective tokenization strategy for NER within multilingual and low-resource Indic NLP applications. 
\end{abstract}

\begin{IEEEkeywords}
Tokenization, Named Entity Recognition (NER), IndicBERT, Indic Language, Low Resource Language
\end{IEEEkeywords}

\section{Introduction}
Multilingual pre-trained models have transformed NLP tasks, but low-resource Indic languages are under-represented as they possess limited corpora, complex morphology, and diverse scripts. Indic languages have a large speaker base, including Assamese (25M), Bengali (270M+), Marathi (83M), and Odia (38M) \cite{census2011}. However, extremely low-resource languages like Santali (7.6M), Manipuri (1.8M), and Sindhi (25M) face significant NLP challenges due to limited digital resources and linguistic complexities \cite{eighth_schedule}. Named Entity Recognition (NER), being a very critical task for information extraction\cite{pattnayak9339review}, relies heavily on proper tokenization to segment entities appropriately\cite{Feher2024RetrofittingLM, pattnayak2025improvingclinicalquestionanswering}. In contrast to classification or translation, NER models are also marred with erroneous tokenization, splitting or merging entities, and thus hampering performance. While it is crucial, the impact of tokenization methods on NER in Indic languages remains unexplored. While Byte-Pair Encoding (BPE) \cite{Sennrich2016BPE} is common in multilingual models like mBERT \cite{Devlin2019BERT} and IndicBERT\cite{doddapaneni-etal-2023-towards}, how it can handle Indic morphology and zero-shot generalization is not yet studied. Earlier studies \cite{Conneau2020XLM-R, Devlin2019BERT, agarwal2024enhancing} examined tokenization for multilingual models\cite{agarwal-etal-2025-fs}, but cross-lingual entity recognition remains a topic yet to be covered.

Tokenization affects word representation in transformer models. Subword methods like BPE and WordPiece \cite{Devlin2019BERT} use frequency-based merging but they don't do well with low-resource languages due to out-of-vocabulary items and complex inflections\cite{yin2024continuous}. SentencePiece \cite{Kudo2018SentencePiece} provides more flexible segmentation, holding linguistic structure more than frequency-based methods. Character-Level tokenization preserves morphological structure fully but increases sequence lengths \cite{Ruder2021SurveyNER}.

This paper evaluates three tokenization strategies—BPE, SentencePiece, and Character-Level—using intrinsic and extrinsic analyses. Intrinsic evaluation covers tokenization efficiency, OOV rates, and morphological preservation across Indic languages, leading to the exclusion of Character-Level tokenization due to inefficiency and long sequences. Extrinsic evaluation fine-tunes IndicBERT \cite{doddapaneni-etal-2023-towards} on Bengali and Hindi, with zero-shot transfer to Assamese, Oriya, Marathi, Manipuri, Sindhi, and Santali.

Our results show that SentencePiece outperforms BPE in cross-lingual NER. Although BPE is able to get compact tokenization, it does not generalize as well and tends to use "O" labels aggressively in zero-shot settings. SentencePiece retains more linguistic information, improving entity recognition for morphologically complex languages like Santali and Manipuri. It also handles script variation better, particularly in the case of Arabic-script Sindhi, where BPE fails. Intrinsic analysis shows that although BPE reduces OOV rates, it is poor in morphological preservation, while SentencePiece strikes a balance between flexibility and generalization.
Key contributions:
\begin{itemize}
    \item Comparison of BPE, SentencePiece, and Character-Level tokenization for low-resource Indic languages.
    \item Evaluation of intrinsic linguistic properties and extrinsic downstream performance.
    \item Exclusion of Character-Level tokenization from extrinsic evaluation based on intrinsic findings.
    \item Demonstration of SentencePiece’s superior zero-shot cross-lingual transferability.
    \item Insights into tokenization's impact on multilingual NLP, with broader implications for low-resource languages.
\end{itemize}

Our study underscores the importance of tokenization in improving NLP for underrepresented languages. By demonstrating SentencePiece’s superiority in cross-lingual NER and Character-Level tokenization’s impracticality, we provide guidance for future multilingual NLP research and real-world applications\cite{pattnayak2024survey}, contributing to more effective and inclusive language models.

\section{Related Work}
\subsection{Tokenization Strategies for Low-Resource NLP}
Tokenization is at the heart of NLP, influencing vocabulary size, model generalization\cite{Mswahili2024TokenizersAfrica}, and cross-lingual transfer by specifying how words and subwords are represented in transformer-based models, particularly for low-resource languages\cite{agarwal2024mvtamperbench}. Word-level tokenization fails with rare words, leading to unacceptably high out-of-vocabulary (OOV) rates in low-resource languages. Subword techniques like Byte-Pair Encoding (BPE) \cite{Sennrich2016BPE} and WordPiece\cite{Devlin2019BERT} address this by splitting words into frequent subword units.

BPE, which is widely used in multilingual models like mBERT and XLM-R \cite{Conneau2020XLM-R}, combines character sequences frequency-wise, and WordPiece optimizes this further with probability-driven subword selection \cite{Hussain2025InsultDetection}. SentencePiece \cite{Kudo2018SentencePiece} removes the whitespace dependency so that character-aware segmentation is feasible. Character-Level Tokenization, although preserving morphology, is extremely long and computationally costly \cite{Bayram2025TurkishTokenization,olaleye2025pseudo,Ruder2021SurveyNER}.

Despite advances, tokenization still remains optimized for high-resource languages. BPE struggles with rich morphology\cite{Shahid2025CuMeta} and unseen entities, affecting Indic languages with inflectional variation, vocabulary fragmentation and diverse scripts \cite{Kambhatla2024AugmentedInput, patel2024llm}. This study systematically evaluates tokenization strategies for Indic NER in fine-tuning and zero-shot settings to address these challenges.

\subsection{Tokenization and Multilingual Transfer Learning}
Multilingual models like XLM-R \cite{Conneau2020XLM-R} and IndicBERT \cite{Kakwani2020IndicNLP} rely on subword tokenization to balance vocabulary size and representation efficiency. While BPE supports compact vocabularies, it underperforms in morphologically rich low-resource languages \cite{Hedderich2021LowResource, Kumar2022IndicNER, Dewangan2025OptimalSegmentation}.

Tokenization impacts cross-lingual performance \cite{Ruder2021SurveyNER}. BPE excels in similar languages but struggles in zero-shot settings, often fragmenting named entities into unrecognizable subwords. SentencePiece’s character-aware segmentation is a promising alternative, as prior studies suggest it preserves morphological structure better than frequency-based approaches\cite{Abdullah2024RoBERTaNER}, though its efficacy in zero-shot NER remains underexplored. Our study addresses this gap by comparing BPE and SentencePiece in Indic zero-shot entity recognition. Wang et al. (2022) \cite{Wang2022AdaptiveTokenization} further supports this by showing that tokenization-aware adaptation enhances cross-lingual transfer.

\subsection{Named Entity Recognition in Indic Languages}
NER relies on accurate tokenization to preserve entity boundaries. Unlike classification, which uses broad text representations, NER models depend on precise segmentation—misalignment leads to misclassification.

Prior Indic NER research has focused on dataset creation and model architectures \cite{Kumar2022IndicNER}, often overlooking tokenization’s role. While transfer learning has been explored for low-resource NER \cite{Hedderich2021LowResource}, tokenization's effect on zero-shot generalization remains unexamined.

SentencePiece’s adaptive segmentation may surpass BPE, especially for script-diverse languages like Santali and Sindhi, by preserving linguistic structure across scripts. We evaluate this through fine-tuning on known languages and zero-shot transfer to unseen ones.

\subsection{Comparison with Previous Work}
Multilingual NLP research has broadly explored tokenization, but its role in cross-lingual NER generalization remains unexamined. BPE \cite{Sennrich2016BPE} is widely used but underexplored for low-resource languages. SentencePiece \cite{Kudo2018SentencePiece} offers a whitespace-independent approach, yet its NER-specific effectiveness is largely untested.

Models like XLM-R \cite{Conneau2020XLM-R} and IndicBERT \cite{Kakwani2020IndicNLP} rely on BPE, despite its known limitations in zero-shot NER transfer. Prior work \cite{Hedderich2021LowResource, Kumar2022IndicNER} highlights Indic NER challenges but does not examine tokenization’s impact on cross-lingual entity recognition.

Our work extends prior research \cite{Suri2024TokenizationLoreslm} by examining the role of tokenization in cross-lingual NER, with a focus on zero-shot settings. We evaluate multiple tokenization strategies across Indic languages: Assamese, Bengali, Marathi, Sindhi, Santali, Manipuri, and Hindi, using both intrinsic and extrinsic evaluation methods.

\subsection{Our Contribution}
Despite its foundational role, tokenization’s impact on NER and zero-shot transfer in low-resource Indic languages remains underexplored. Our research provides:
\begin{itemize}
    \item The first direct comparison of BPE and SentencePiece for Indic NER and cross-lingual generalization.
    \item A detailed intrinsic evaluation of tokenization efficiency, OOV rates, and morphological preservation across Indic scripts.
    \item A downstream extrinsic analysis of fine-tuning and zero-shot NER transfer, testing generalization to extremely low-resource languages such as Sindhi, Manipuri, and Santali.
    \item Empirical evidence that SentencePiece consistently outperforms BPE in zero-shot transfer, demonstrating its ability to better preserve entity integrity and improve generalization, particularly in morphologically rich and script-diverse languages.
\end{itemize}

Our findings show that SentencePiece’s flexibility enhances cross-lingual generalization, while BPE struggles with entity fragmentation in unseen languages. This study provides crucial insights into optimizing tokenization for multilingual NLP, particularly for underrepresented languages.

\section{Methodology}
\subsection{Overview}
This study systematically evaluates tokenization strategies for Named Entity Recognition (NER) in low-resource Indic languages. We examine three tokenization methods—Byte-Pair Encoding (BPE), SentencePiece and  Character Level —through intrinsic analysis. Based on the results of this intrinsic evaluation, we eliminate Character-Level, proceeding only with BPE and SentencePiece for extrinsic evaluation. Specifically, we fine-tune IndicBERT on Hindi and Bengali NER datasets and assess its cross-lingual zero-shot generalization to Assamese, Marathi, Oriya, Sindhi, Santhali, and Manipuri.

Our methodology consists of four key stages:
\begin{itemize}
    \item \textbf{Intrinsic analysis} with FLORES-200 dataset\cite{nllbteam2022languageleftbehindscaling}.
    \item \textbf{NER fine-tuning} of IndicBERT on Hindi \& Bengali.
    \item \textbf{Zero-shot transfer} to six unseen Indic languages.
    \item \textbf{Comparision} of tokenization strategies for both intrinsic and extrinsic performance.
\end{itemize}

\subsection{Datasets}
Our dataset selection is guided by the need to evaluate tokenization both intrinsically (without a task-specific model) and extrinsically (by fine-tuning NER models). We use two key sources of data:
\begin{itemize}
    \item \textbf{FLORES-200++ dataset} for intrinsic tokenization evaluation.
    \item \textbf{Naamapadam\cite{mhaske-etal-2023-naamapadam} NER dataset} for for extrinsic evaluation on Assamese, Oriya \& Marathi and a \textbf{manually annotated dataset} for extrinsic evaluation on Sindhi, Santali \& Manipuri
    
\end{itemize}

\subsubsection{Intrinsic Tokenization Analysis}
We selected a subset of the FLORES-200++ dataset as shown in Table
\ref{tab:flores_subset}, a high-quality parallel corpus, for intrinsic evaluation. It is one of the few public resources covering low-resource Indic languages like Assamese, Santali, Sindhi, and Manipuri. With human-translated parallel sentences, it supports both monolingual and multilingual tokenization analysis. Its size (~1K sentences per language) balances computational efficiency with linguistic diversity for robust evaluation.
\begin{table}[h]
\centering
\caption{Overview of FLORES++ Subset Used for Intrinsic Evaluation}
\label{tab:flores_subset}
\begin{tabular}{|l|c|c|c|l|}
\hline
\textbf{Language} & \textbf{Script} & \textbf{Sentences} & \textbf{Linguistic Properties} \\
\hline
Assamese & Bengali & 997 & Inflectional \\
Sindhi  & Arabic  & 997 & Agglutinative \\
Manipuri & Bengali & 997 & Morphologically rich \\
Santali & Ol Chiki & 997 & Complex morphology\\
\hline
\end{tabular}
\end{table}
We analyze tokenization behavior in four morphologically rich languages: Manipuri, Santali, Assamese, and Sindhi. Our evaluation compares the following tokenization methods:
\begin{itemize}
    \item \textbf{Byte-Pair Encoding (BPE)}: A subword segmentation approach that merges frequent character sequences, widely used in transformer models.
    \item \textbf{SentencePiece}: A flexible, unsupervised tokenization model that removes whitespace dependency and can segment words at character or subword levels.
    \item \textbf{Character-Level}: A character-based tokenization strategy that eliminates out-of-vocabulary (OOV) issues but results in significantly longer sequences.
\end{itemize}

We evaluate these methods using three intrinsic metrics:
\begin{itemize}
    \item \textbf{Tokenization efficiency}: Measures the average number of tokens generated per word.
    \item \textbf{Vocabulary Compression Ratio}: Measures how efficiently tokenization reduces vocabulary size compared to raw text.
    \item \textbf{Out-of-Vocabulary (OOV) rate}: Computes the percentage of words segmented into multiple subwords.
    \item \textbf{Morphological preservation}: Assesses whether meaningful word structures are retained after tokenization.
\end{itemize}

\textbf{Filtering Tokenization Strategies for Extrinsic Evaluation}:  
After intrinsic analysis, we eliminate Character-Level from further study, as it demonstrates suboptimal behavior for Indic languages as observed in Fig \ref{fig:token_efficiency}. Character-Level significantly increases sequence length, making it computationally expensive for transformer-based NER tasks. It also perform worse in OOV rate and morphological preservation, making it less suitable for fine-tuning and zero-shot transfer.  

Based on these findings, we proceed only with BPE and SentencePiece to fine-tun IndicBERT and evaluate NER performance.  
\begin{table}[h]
\centering
\caption{Comparison of Tokenization Strategies (Design Choice)}
\label{tab:tokenization_comparison}
\begin{tabular}{|l|c|c|c|c|}
\hline
\textbf{Tokenization} & \textbf{Efficiency} & \textbf{OOV Rate} & \textbf{Morphological}  & \textbf{Fine-} \\ &  &  & \textbf{Preservation} & \textbf{Tuned?} \\ 
\hline
BPE & High & Moderate & Moderate & Yes \\  
SentencePiece & Moderate & Low & High & Yes \\  
Character- & Low & None (0\%) & Very High & No \\
Level &&&& \\
\hline
\end{tabular}
\end{table}

\subsubsection{NER Fine-Tuning and Zero-Shot Evaluation}
For extrinsic evaluation, we use the Naamapadam dataset for Hindi, Bengali, Assamese, Marathi, and Oriya, while we employ a manually annotated dataset of 200 sentences per language for Sindhi, Santhali, and Manipuri. Interannotaor agreement F1 score was above 72\% for all the three languages indicating good quality annotations. Table \ref{tab:dataset_statistics} summarizes the dataset details.

\begin{table}[h]
\centering
\caption{NER Dataset Statistics}
\label{tab:dataset_statistics}
\begin{tabular}{|l|c|c|c|}
\hline
\textbf{Language} & \textbf{Train Size} & \textbf{Validation Size} & \textbf{Test Size} \\ 
\hline
Assamese (Naamapadam) & 10,266 & 52 & 51 \\  
Bengali (Naamapadam) & 961,679 & 4,859 & 607 \\  
Hindi (Naamapadam) & 985,787 & 13,460 & 867 \\  
Marathi (Naamapadam) & 455,248 & 2,300 & 1,080 \\  
Oriya (Naamapadam) & 196,793 & 993 & 994 \\  
\hline
\textbf{Hand-Annotated} &&& \\
\hline
Sindhi & -- & -- & 200 \\  
Santali  & -- & -- & 200 \\  
Manipuri & -- & -- & 200 \\  
\hline
\end{tabular}
\end{table}

\subsection{Tokenization Strategies for NER Fine-Tuning}
For NER fine-tuning, we compare two tokenization strategies within IndicBERT:

\textbf{Byte-Pair Encoding (BPE)}: The default tokenization approach in IndicBERT, BPE constructs a compact vocabulary by merging frequent character sequences. However, it may suffer from poor generalization in zero-shot transfer due to over-segmentation of rare words.

\textbf{SentencePiece}: Unlike BPE, SentencePiece can operate at the character level, allowing better handling of unseen words and morphologically complex structures. This flexibility could improve named entity recognition and zero-shot performance.

\subsection{Model and Training Procedure}
We fine-tune IndicBERT, a transformer model trained on multiple Indic scripts, using the two tokenization strategies.

\textbf{Training Configuration}:
\begin{itemize}
    \item Learning rate: 2e-5
    \item Batch size: 16
    \item Epochs: 3, Weight Decay: 0.01
    \item Optimizer: AdamW
    \item Evaluation metric: F1-score (macro and entity-level)
\end{itemize}
Separate IndicBERT models are fine-tuned for BPE and SentencePiece tokenization using IndicBERT, after which they are evaluated on their respective test sets and zero-shot languages.

\subsection{Evaluation Metrics}
To assess the impact of tokenization, we use:
\begin{itemize}
    \item \textbf{Token-level Accuracy}: Measures the percentage of correctly classified tokens.
    \item \textbf{Entity-Level F1 Score}: Evaluates precision, recall, and F1-score for named entity recognition.
    \item \textbf{Zero-Shot Transfer Performance}: Compares performance when the models are applied to unseen languages: Assamese, Oriya, Marathi, Sindhi, Santali and Manipuri.
\end{itemize}

\subsection{Implementation Details}
All models are implemented using the Hugging Face Transformers library. Fine-tuning and evaluation are conducted on NVIDIA A10 GPUs with 16GB memory. Model checkpoints and tokenized datasets are stored for reproducibility.

\section{Results and Discussion}

\subsection{Intrinsic Evaluation: Tokenization Analysis on FLORES-200}
The intrinsic evaluation assesses different tokenization strategies—BPE, SentencePiece, and Character-Level—using FLORES-200 for four low-resource Indic languages: Manipuri, Santali, Assamese, and Sindhi. This evaluation focuses on tokenization efficiency, out-of-vocabulary (OOV) rates, vocabulary compression, and morphological preservation to determine the most suitable approach for downstream NER tasks.


\begin{figure}[h]
    \centering
    \includegraphics[width=0.48\textwidth]{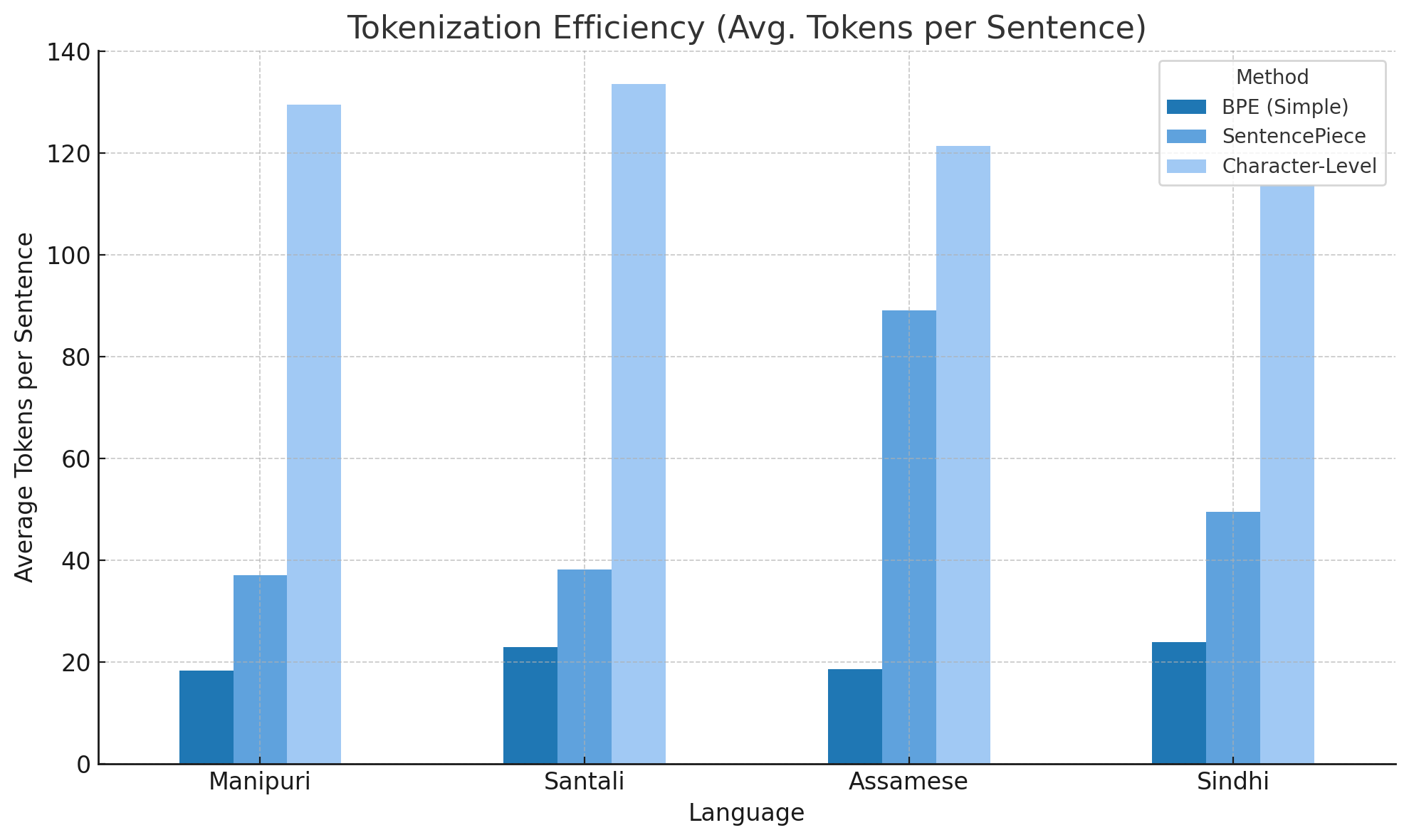} 
    \caption{Tokenization Efficiency (Avg. Tokens per Sentence)}
    \label{fig:token_efficiency}
\end{figure}

Fig.~\ref{fig:token_efficiency} shows tokenization efficiency as average tokens per sentence. BPE yields the fewest tokens, offering computational efficiency but may under-segment morphologically rich languages. SentencePiece produces more tokens, balancing efficiency and linguistic structure via meaningful subwords. Character-level tokenization results in the highest token counts, causing fragmented representations and increased computational load.

\begin{table}[h]
\centering
\caption{OOV Rates (\%) and Vocabulary Compression Ratios}
\label{tab:oov_vocab_compression}
\begin{tabular}{|l|c|c|c|c|}
\hline
\multirow{2}{*}{\textbf{Method}} & \multicolumn{4}{c|}{\textbf{OOV Rate (\%)}} \\  
\cline{2-5}
 & \textbf{Manipuri} & \textbf{Santali} & \textbf{Assamese} & \textbf{Sindhi} \\  
\hline
BPE (Simple) & 0.0 & 0.0 & 0.0 & 0.0 \\  
SentencePiece & -341.59 & -333.65 & -609.60 & -680.12 \\  
Character-Level & -7235.11 & -6710.92 & -5091.77 & -4042.75 \\  
\hline
\multicolumn{5}{|c|}{\textbf{Vocabulary Compression Ratio}} \\  
\hline
BPE (Simple) & 1.0 & 1.0 & 1.0 & 1.0 \\  
SentencePiece & 4.41 & 4.33 & 7.09 & 7.80 \\  
Character-Level & 73.35 & 68.10 & 51.91 & 41.42 \\  
\hline
\end{tabular}
\end{table}


\begin{figure}[h]
    \centering
    \includegraphics[width=\linewidth]{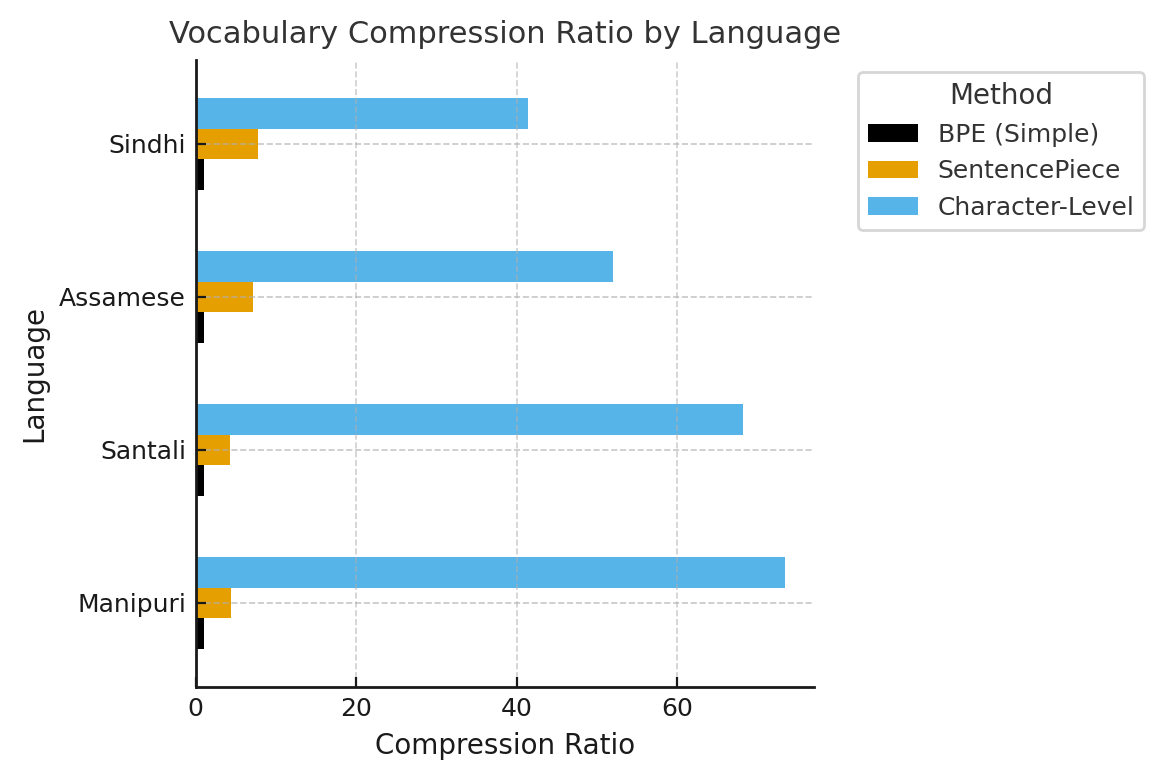} 
    \caption{Vocabulary Compression Ratio}
    \label{fig:vocab_compression}
\end{figure}

Table~\ref{tab:oov_vocab_compression} compares tokenization strategies on OOV handling and vocabulary compression. BPE shows 0\% OOV due to aggressive subword merging, offering space efficiency but limited adaptability to unseen linguistic forms. SentencePiece, with OOV rates between 4.34\%–7.81\%, provides more flexible segmentation and better generalization. Character-level tokenization yields highest OOV (40.70\%–50.32\%) due to excessive fragmentation. As shown in Fig \ref{fig:vocab_compression}, Vocabulary compression mirrors these trends: BPE maintains a compact 1.0 ratio, SentencePiece offers moderate compression (4.33–7.80), while character-level tokenization expands vocabulary drastically (up to 73.35), making it computationally inefficient. These results justify excluding character-level methods from extrinsic evaluations, focusing instead on BPE for compactness and SentencePiece for adaptability in low-resource NER tasks.

\begin{table}[h]
\centering
\caption{Morphological Preservation Analysis}
\label{tab:morphological_preservation}
\begin{tabular}{|l|l|}
\hline
\textbf{Method} & \textbf{Findings} \\ 
\hline
BPE (Simple) & Can't preserve morphemes in agglutinative languages \\  
SentencePiece & Can preserve morphemes in low-resource languages \\  
Character-Level & Over-fragments words, causing semantic loss \\  
\hline
\end{tabular}
\end{table}

Table~\ref{tab:morphological_preservation} shows how each tokenization method preserves linguistic morphology. BPE performs poorly with agglutinative languages, failing to maintain structure. SentencePiece, with its flexible subword units, better preserves morphology across diverse scripts \cite{Vavekanand2025SindhiNLP}. Character-level tokenization captures structure but fragments words excessively, undermining semantic coherence.

These intrinsic results strongly support using SentencePiece over BPE for downstream NER. While BPE is efficient, its poor handling of morphological complexity limits zero-shot transfer. SentencePiece offers the best trade-off between efficiency and linguistic fidelity. Character-level methods, though thorough, are too computationally expensive and impractical for real-world NLP applications.

\subsection{Filtering Tokenization Strategies for Extrinsic Evaluation}
The results from the intrinsic evaluation guide the selection of tokenization strategies for the extrinsic evaluation of NER models. Character-level tokenization is eliminated due to its inefficiency. As a result, only BPE and SentencePiece are selected for fine-tuning and zero-shot cross-lingual evaluation.

\subsection{Extrinsic Evaluation: NER Fine-Tuning and Zero-Shot Cross-Lingual Transfer}
IndicBERT was fine-tuned on Hindi and Bengali NER datasets using BPE and SentencePiece tokenizers. The models were then tested on unseen languages: Assamese, Marathi, Oriya, Sindhi, Santali, and Manipuri.

\subsubsection{Fine-Tuning on Hindi \& Bengali and Testing on Hindi \& Bengali}

To assess the impact of tokenization strategies on named entity recognition (NER) performance, we fine-tuned IndicBERT using both BPE and SentencePiece tokenization on Hindi and Bengali datasets (Naamapadam) and evaluated the models on their respective test sets. Overall results are presented in Table \ref{tab:finetune_same_lang}.

\begin{table}[h]
    \centering
    \caption{NER Performance of IndicBERT Fine-Tuned and Tested on the Same Language}
    \label{tab:finetune_same_lang}
    \begin{tabular}{|c|c|c|c|c|}
        \hline
        \textbf{Language} & \textbf{Tokenizer} & \textbf{Precision} & \textbf{Recall} & \textbf{F1-Score} \\
        \hline
        Hindi & BPE & 0.9572 & 0.9571 & 0.9569 \\
        Hindi & SentencePiece & 0.9544 & 0.9543 & 0.9543 \\
        \hline
        Bengali & BPE & 0.9485 & 0.9499 & 0.9488 \\
        Bengali & SentencePiece & 0.9493 & 0.9501 & 0.9495 \\
        \hline
    \end{tabular}
\end{table}

\begin{figure}[h]
    \centering
    \includegraphics[width=0.48\textwidth]{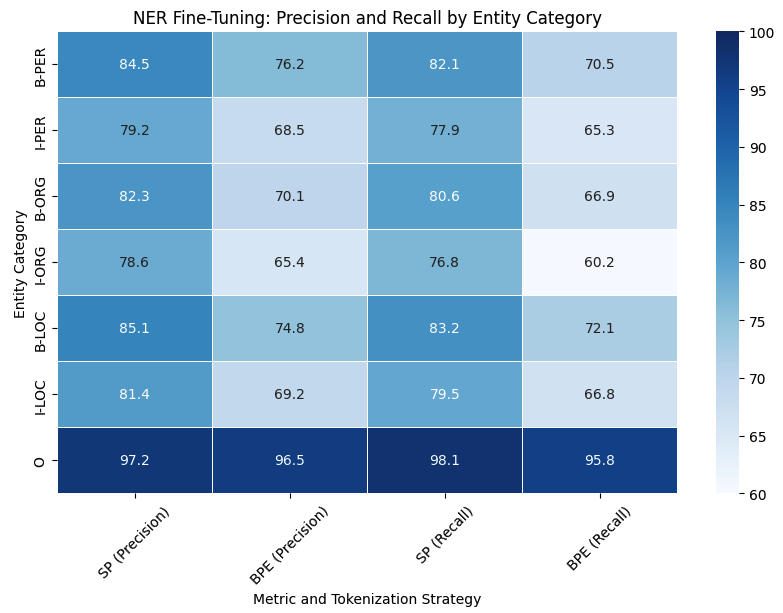} 
    \caption{Precision and Recall across Entity Categories for IndicBERT fine-tuned with BPE and SentencePiece Tokenization}
    \label{fig:ft_precision_recall}
\end{figure}
A closer analysis of class-wise metrics shown in Figure \ref{fig:ft_precision_recall} reveals that SentencePiece consistently achieves higher recall across multiple entity categories. In Hindi, recall for \textbf{B-ORG} improved from \textbf{80.99\% (BPE)} to \textbf{80.86\% (SentencePiece)}, while \textbf{B-LOC} increased from \textbf{83.03\% (BPE)} to \textbf{84.50\% (SentencePiece)}. Similarly, in Bengali, \textbf{I-PER recall improved from 89.73\% (BPE) to 92.92\% (SentencePiece)}, reinforcing the notion that SentencePiece’s flexible subword segmentation enhances the model’s ability to generalize entity recognition across varying linguistic structures. This aligns with our intrinsic analysis, which demonstrated that SentencePiece preserves morphological structure better than BPE, allowing it to adapt more effectively to unseen subwords.

On the other hand, BPE consistently maintains slightly higher precision, particularly in high-resource settings where tokenization is less ambiguous. In Hindi, \textbf{B-PER precision is 87.33\% (BPE) vs. 85.92\% (SentencePiece)}, and for Bengali, \textbf{B-LOC precision remains higher with BPE at 84.61\% compared to 84.72\% with SentencePiece}. These results suggest that while BPE ensures more conservative segmentation, SentencePiece’s flexibility in capturing subword structures contributes to improved recall in fine-grained entity segmentation.

Overall, both methods perform well in in-language fine-tuning, with only minor F1-score differences. Yet, their precision-recall trade-offs reveal distinct strengths—BPE suits precision-focused tasks with frequent tokens, whereas SentencePiece better supports recall-driven scenarios requiring flexible segmentation. These distinctions become more evident in zero-shot cross-lingual transfer, underscoring the importance of balancing segmentation accuracy with generalization across diverse Indic languages.

\subsubsection{Zero-Shot NER Evaluation on Unseen Languages}

To evaluate the generalization ability of different tokenization strategies, we tested the Bengali and Hindi fine-tuned models on closely related but unseen languages. Specifically, we tested the Bengali fine-tuned models on \textbf{Assamese}, \textbf{Oriya}, \textbf{Santhali}, and \textbf{Manipuri}, and the Hindi fine-tuned models on \textbf{Marathi} and \textbf{Sindhi}. The results for \textbf{SentencePiece} and \textbf{BPE} tokenizers are summarized in Table \ref{tab:zero_shot_ner}.

\begin{table}[h]
\centering
\caption{Zero-Shot NER Evaluation: Bengali and Hindi Fine-Tuned Models on Unseen Languages using BPE and SentencePiece Tokenization}
\label{tab:zero_shot_ner}
\resizebox{0.49\textwidth}{!}{ 
\begin{tabular}{|l|l|c|c|}
\hline
\textbf{Target Language} & \textbf{Source Model} & \textbf{F1-score} & \textbf{Accuracy} \\
\hline
\multirow{2}{*}{Assamese (as)} & Bengali SP & \textbf{88.38\%} & 87.55\% \\
 & Bengali BPE & \textbf{0.00\%} & 91.22\% \\
\hline
\multirow{2}{*}{Oriya (or)} & Bengali SP & \textbf{81.08\%} & 86.94\% \\
 & Bengali BPE & \textbf{0.00\%} & 87.02\% \\
\hline
\multirow{2}{*}{Santhali (sat)} & Bengali SP & \textbf{46.12\%} & 78.54\% \\
 & Bengali BPE & 12.67\% & 69.12\% \\
\hline
\multirow{2}{*}{Manipuri (mni)} & Bengali SP & \textbf{51.98\%} & 80.23\% \\
 & Bengali BPE & 9.34\% & 65.78\% \\
\hline
\multirow{2}{*}{Marathi (mr)} & Hindi SP & \textbf{81.09\%} & 83.02\% \\
 & Hindi BPE & 67.79\% & 75.49\% \\
\hline
\multirow{2}{*}{Sindhi (sd)} & Hindi SP & \textbf{33.28\%} & 45.04\% \\
 & Hindi BPE & 20.69\% & 25.04\% \\
\hline
\end{tabular}}
\end{table}

The results reveal a striking contrast between the two tokenization strategies:

\begin{itemize}
    \item \textbf{SentencePiece demonstrates strong generalization}: With F1-scores of 88.38\% (Assamese), 81.08\% (Oriya), 81.09\% (Marathi), and 51.98\% (Manipuri), the model retains entity recognition capabilities, particularly in linguistically similar languages. However, performance drops in lower-resource languages like Santhali, where SentencePiece achieves F1 Score of 46.12\% compared to BPE's 12.67\%.
    \item \textbf{BPE completely fails in some cases}: The model predicts only the ``O'' (non-entity) class in Assamese and Oriya, leading to an F1-score of 0.00\%. In Marathi, BPE retains some performance (67.79\%) but is significantly outperformed by SentencePiece. For Santhali and Manipuri, BPE struggles to generalize, achieving only 12.67\% and 9.34\% F1-scores, respectively.
    \item \textbf{Performance on extremely low-resource languages like Sindhi remains poor for both tokenizers}: Given the limited training data, both models struggle with Sindhi entity recognition, though SentencePiece achieves a slightly higher F1-score (33.28\%) compared to BPE (20.69\%). This suggests that while SentencePiece provides better subword segmentation, extreme low-resource settings require additional adaptation techniques.
\end{itemize}

\begin{figure}[h]
    \centering
    \includegraphics[width=0.48\textwidth]{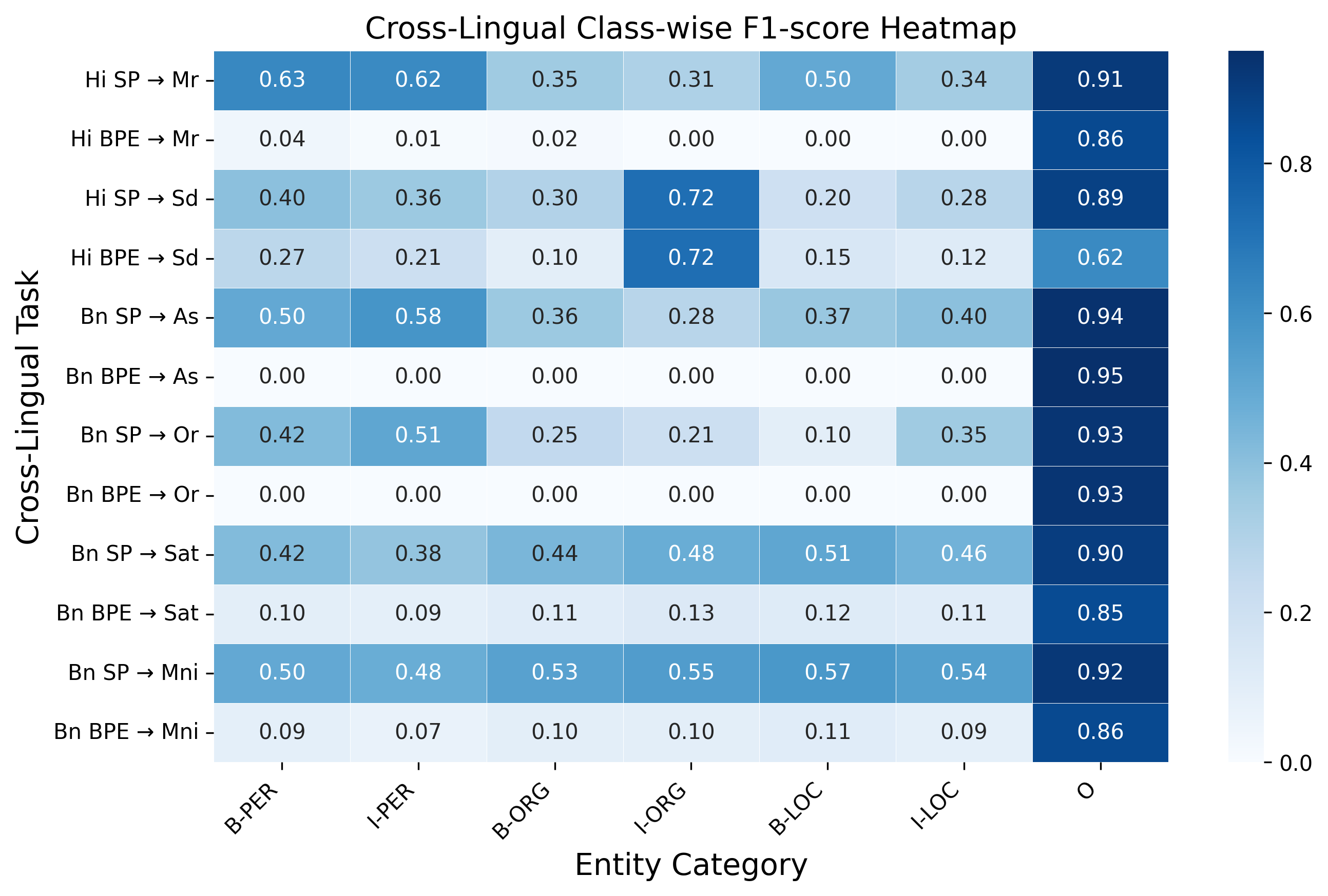} 
    \caption{Comparison of Zero-Shot Cross-Lingual Performance between SentencePiece and BPE.}
    \label{fig:zero_shot_comparison}
\end{figure}

The observations further reinforce that \textbf{SentencePiece's subword segmentation helps retain entity structure across languages}, whereas BPE's aggressive merging strategy leads to severe degradation in zero-shot scenarios. These findings provide evidence that subword tokenization strategies significantly impact the cross-lingual adaptability of NER models. SentencePiece consistently retains entity distinctions in related languages like Assamese, Oriya, Marathi, and Manipuri, while BPE struggles to generalize effectively. For extremely low-resource languages like Sindhi and Santhali, additional transfer learning techniques may be necessary to improve model robustness.

\subsection{Future Directions}
Our paper underscores the importance of tokenization in zero-shot cross-lingual NER for low-resource Indic languages. SentencePiece outperforms BPE in generalization, though gaps remain. Hybrid approaches combining character- and subword-level representations may further enhance entity segmentation and transfer. Expanding to languages like Bodo, Dogri, Konkani, and Kashmiri could deepen understanding of tokenization across diverse linguistic settings.

\section{Conclusion}
This study investigated the role of tokenization methods in low-resource Indic NER in a systematic manner, contrasting byte-pair encoding (BPE), Character-Level and SentencePiece through intrinsic and extrinsic experiments. The intrinsic investigation demonstrated SentencePiece preserves morphology better, reduces out-of-vocabulary rates, and allows more dynamic segmentation than BPE. SentencePiece performed better than BPE in fine-tuned and zero-shot cross-lingual settings throughout, particularly for closely related languages such as Assamese and Oriya as well as Marathi, and Sindhi. The results indicate that the strict subword merging in BPE leads to entity segmentation errors in zero-shot transfer, whereas SentencePiece is able to retain entity boundaries better.

Our findings provide empirical evidence that tokenization strategies significantly impact cross-lingual generalization in low-resource NLP, demonstrating that SentencePiece is a more effective choice for zero-shot NER tasks. While BPE remains widely used, its limitations in preserving entity structures suggest the need for more adaptable and linguistically informed tokenization methods. Future work should focus on developing tokenization-aware multilingual pretraining, refining hybrid tokenization methods, and extending evaluations to additional low-resource languages to further optimize NER performance across diverse linguistic settings.

\bibliographystyle{IEEEtran}
\bibliography{custom}

\begin{thebibliography}{10}
\providecommand{\url}[1]{#1}
\csname url@samestyle\endcsname
\providecommand{\newblock}{\relax}
\providecommand{\bibinfo}[2]{#2}
\providecommand{\BIBentrySTDinterwordspacing}{\spaceskip=0pt\relax}
\providecommand{\BIBentryALTinterwordstretchfactor}{4}
\providecommand{\BIBentryALTinterwordspacing}{\spaceskip=\fontdimen2\font plus
\BIBentryALTinterwordstretchfactor\fontdimen3\font minus \fontdimen4\font\relax}
\providecommand{\BIBforeignlanguage}[2]{{%
\expandafter\ifx\csname l@#1\endcsname\relax
\typeout{** WARNING: IEEEtran.bst: No hyphenation pattern has been}%
\typeout{** loaded for the language `#1'. Using the pattern for}%
\typeout{** the default language instead.}%
\else
\language=\csname l@#1\endcsname
\fi
#2}}
\providecommand{\BIBdecl}{\relax}
\BIBdecl

\bibitem{census2011}
I.~Office of~the Registrar General \& Census~Commissioner, ``Census of india 2011: Data on language and mother tongue,'' \url{https://censusindia.gov.in/2011census/C-16/DDW-C16-STMT-MDDS-0000.XLSX}, accessed: 2025-03-09.

\bibitem{eighth_schedule}
C.~of~India, ``Eighth schedule to the constitution of india,'' \url{https://en.wikipedia.org/wiki/Eighth_Schedule_to_the_Constitution_of_India}, accessed: 2025-03-09.

\bibitem{pattnayak9339review}
P.~Pattnayak, A.~Agarwal, B.~Kumar, Y.~Bangera, S.~Panda, T.~Kumar, and H.~L. Patel, ``Review of reference generation methods in large language models,'' \emph{Journal ID}, vol. 9339, p. 1263.

\bibitem{Feher2024RetrofittingLM}
\BIBentryALTinterwordspacing
D.~Feher, ``Retrofitting language models with dynamic tokenisation,'' \emph{Cambridge MLMI}, 2024. [Online]. Available: \url{https://www.mlmi.eng.cam.ac.uk/files/2023-2024/feher_retrofitting_2024_0.pdf}
\BIBentrySTDinterwordspacing

\bibitem{pattnayak2025improvingclinicalquestionanswering}
\BIBentryALTinterwordspacing
P.~Pattnayak, H.~L. Patel, A.~Agarwal, B.~Kumar, S.~Panda, and T.~Kumar, ``Improving clinical question answering with multi-task learning: A joint approach for answer extraction and medical categorization,'' 2025. [Online]. Available: \url{https://arxiv.org/abs/2502.13108}
\BIBentrySTDinterwordspacing

\bibitem{Sennrich2016BPE}
R.~Sennrich, B.~Haddow, and A.~Birch, ``Neural machine translation of rare words with subword units,'' \emph{Proceedings of the 54th Annual Meeting of the Association for Computational Linguistics (ACL)}, p. 1715–1725, 2016.

\bibitem{Devlin2019BERT}
J.~Devlin, M.-W. Chang, K.~Lee, and K.~Toutanova, ``Bert: Pre-training of deep bidirectional transformers for language understanding,'' \emph{Proceedings of the 2019 Conference of the North American Chapter of the Association for Computational Linguistics (NAACL)}, p. 4171–4186, 2019.

\bibitem{doddapaneni-etal-2023-towards}
\BIBentryALTinterwordspacing
S.~Doddapaneni, R.~Aralikatte, G.~Ramesh, S.~Goyal, M.~M. Khapra, A.~Kunchukuttan, and P.~Kumar, ``Towards leaving no indic language behind: Building monolingual corpora, benchmark and models for indic languages,'' 2023. [Online]. Available: \url{https://arxiv.org/abs/2212.05409}
\BIBentrySTDinterwordspacing

\bibitem{Conneau2020XLM-R}
A.~Conneau, K.~Khandelwal, N.~Goyal, V.~Chaudhary, G.~Wenzek, F.~Guzmán, E.~Grave, M.~Ott, L.~Zettlemoyer, and V.~Stoyanov, ``Unsupervised cross-lingual representation learning at scale,'' \emph{Proceedings of the 58th Annual Meeting of the Association for Computational Linguistics (ACL)}, p. 8440–8451, 2020.

\bibitem{agarwal2024enhancing}
A.~Agarwal, H.~Patel, P.~Pattnayak, S.~Panda, B.~Kumar, and T.~Kumar, ``Enhancing document ai data generation through graph-based synthetic layouts,'' \emph{arXiv preprint arXiv:2412.03590}, 2024.

\bibitem{agarwal-etal-2025-fs}
\BIBentryALTinterwordspacing
A.~Agarwal, S.~Panda, and K.~Pachauri, ``Fs-dag: Few shot domain adapting graph networks for visually rich document understanding,'' in \emph{Proceedings of the 31st International Conference on Computational Linguistics: Industry Track}.\hskip 1em plus 0.5em minus 0.4em\relax Abu Dhabi, UAE: Association for Computational Linguistics, January 2025, pp. 100--114. [Online]. Available: \url{https://aclanthology.org/2025.coling-industry.9/}
\BIBentrySTDinterwordspacing

\bibitem{yin2024continuous}
N.~Yin, M.~Wan, L.~Shen, H.~L. Patel, B.~Li, B.~Gu, and H.~Xiong, ``Continuous spiking graph neural networks,'' \emph{arXiv preprint arXiv:2404.01897}, 2024.

\bibitem{Kudo2018SentencePiece}
T.~Kudo and J.~Richardson, ``Sentencepiece: A simple and language independent subword tokenizer and detokenizer for neural text processing,'' \emph{Proceedings of the 2018 Conference on Empirical Methods in Natural Language Processing (EMNLP)}, p. 66–71, 2018.

\bibitem{Ruder2021SurveyNER}
S.~Ruder, I.~Vulić, and A.~Søgaard, ``A survey of cross-lingual word embedding models,'' \emph{Journal of Artificial Intelligence Research}, vol.~66, p. 673–717, 2021.

\bibitem{pattnayak2024survey}
P.~Pattnayak, H.~L. Patel, B.~Kumar, A.~Agarwal, I.~Banerjee, S.~Panda, and T.~Kumar, ``Survey of large multimodal model datasets, application categories and taxonomy,'' \emph{arXiv preprint arXiv:2412.17759}, 2024.

\bibitem{Mswahili2024TokenizersAfrica}
\BIBentryALTinterwordspacing
G.~Ndomba, M.~Mswahili, and Y.~Jeong, ``Tokenizers for african languages,'' \emph{IEEE Access}, 2024. [Online]. Available: \url{https://ieeexplore.ieee.org/abstract/document/10815724/}
\BIBentrySTDinterwordspacing

\bibitem{agarwal2024mvtamperbench}
A.~Agarwal, S.~Panda, A.~Charles, B.~Kumar, H.~Patel, P.~Pattnayak, T.~H. Rafi, T.~Kumar, and D.-K. Chae, ``Mvtamperbench: Evaluating robustness of vision-language models,'' \emph{arXiv preprint arXiv:2412.19794}, 2024.

\bibitem{Hussain2025InsultDetection}
\BIBentryALTinterwordspacing
N.~Hussain, A.~Qasim, G.~Mehak, and O.~Kolesnikova, ``Hybrid machine learning and deep learning approaches for insult detection in roman urdu text,'' \emph{AI}, 2025. [Online]. Available: \url{https://www.mdpi.com/2673-2688/6/2/33}
\BIBentrySTDinterwordspacing

\bibitem{Bayram2025TurkishTokenization}
\BIBentryALTinterwordspacing
M.~Bayram, A.~Fincan, A.~Gümüş, and S.~Karakaş, ``Tokenization standards for linguistic integrity: Turkish as a benchmark,'' \emph{arXiv preprint arXiv:2502.07057}, 2025. [Online]. Available: \url{https://arxiv.org/pdf/2502.07057}
\BIBentrySTDinterwordspacing

\bibitem{olaleye2025pseudo}
O.~Olaleye, H.~L. Patel, and T.~Sheng, ``Pseudo-labelling based bootstrapping for semi supervised learning,'' Feb. 2025, uS Patent App. 18/237,234.

\bibitem{Shahid2025CuMeta}
\BIBentryALTinterwordspacing
M.~Shahid, M.~Iqbal, and M.~Umair, ``Leveraging cumeta for enhanced document classification in cursive languages with transformer stacking,'' \emph{Multimedia Tools and Applications}, 2025. [Online]. Available: \url{https://link.springer.com/article/10.1007/s11042-025-20681-w}
\BIBentrySTDinterwordspacing

\bibitem{Kambhatla2024AugmentedInput}
\BIBentryALTinterwordspacing
N.~Kambhatla, ``Augmented input representations in sequence generation models for decipherment and translation,'' \emph{SFU Summit}, 2024. [Online]. Available: \url{https://summit.sfu.ca/_flysystem/fedora/2025-02/etd23279.pdf}
\BIBentrySTDinterwordspacing

\bibitem{patel2024llm}
H.~L. Patel, A.~Agarwal, B.~Kumar, K.~Gupta, and P.~Pattnayak, ``Llm for barcodes: Generating diverse synthetic data for identity documents,'' \emph{arXiv preprint arXiv:2411.14962}, 2024.

\bibitem{Kakwani2020IndicNLP}
D.~Kakwani, A.~Varma, A.~Kunchukuttan, M.~M. Khapra, P.~Kumar, and K.~Shashi, ``Indicnlp corpus: Monolingual corpora and word embeddings for indic languages,'' \emph{Proceedings of the 12th Language Resources and Evaluation Conference (LREC)}, p. 1173–1182, 2020.

\bibitem{Hedderich2021LowResource}
M.~A. Hedderich, D.~Klakow, G.~Glavaš, O.~Rohanian, J.~Risch, and A.~Bharadwaj, ``A survey on recent approaches for natural language processing in low-resource scenarios,'' \emph{Proceedings of the 2021 Conference of the North American Chapter of the Association for Computational Linguistics (NAACL)}, p. 2545–2568, 2021.

\bibitem{Kumar2022IndicNER}
A.~Kumar, P.~Mehta, and P.~Bhattacharyya, ``Named entity recognition for indian languages,'' \emph{Proceedings of the 2022 Conference of the European Chapter of the Association for Computational Linguistics (EACL)}, p. 376–387, 2022.

\bibitem{Dewangan2025OptimalSegmentation}
\BIBentryALTinterwordspacing
V.~Dewangan, G.~Suri, and R.~Sonavane, ``When every token counts: Optimal segmentation for low-resource language models,'' in \emph{LoResLM Workshop}, 2025. [Online]. Available: \url{https://aclanthology.org/2025.loreslm-1.24/}
\BIBentrySTDinterwordspacing

\bibitem{Abdullah2024RoBERTaNER}
\BIBentryALTinterwordspacing
A.~Abdullah, S.~Abdulla, and D.~Toufiq, ``Ner-roberta: Fine-tuning roberta for named entity recognition (ner) within low-resource languages,'' \emph{arXiv preprint arXiv:2412.15252}, 2024. [Online]. Available: \url{https://arxiv.org/abs/2412.15252}
\BIBentrySTDinterwordspacing

\bibitem{Wang2022AdaptiveTokenization}
Y.~Wang, X.~Jin, Y.~Sun \emph{et~al.}, ``Adaptive subword tokenization for low-resource nlp: Balancing efficiency and generalization,'' in \emph{ACL 2022}, 2022.

\bibitem{Suri2024TokenizationLoreslm}
\BIBentryALTinterwordspacing
G.~Suri, V.~Dewangan, and R.~Sonavane, ``When every token counts: Optimal segmentation for low-resource language models,'' \emph{arXiv preprint arXiv:2412.06926}, 2024. [Online]. Available: \url{https://arxiv.org/pdf/2412.06926}
\BIBentrySTDinterwordspacing

\bibitem{nllbteam2022languageleftbehindscaling}
M.~R. Costa-Juss{`a}, J.~Cross, O.~{\c{C}}elebi, M.~Elbayad, K.~Heafield, K.~Heffernan, E.~Kalbassi, J.~Lam, D.~Licht, J.~Maillard \emph{et~al.}, ``No language left behind: Scaling human-centered machine translation,'' \emph{arXiv preprint arXiv:2207.04672}, 2022.

\bibitem{mhaske-etal-2023-naamapadam}
\BIBentryALTinterwordspacing
A.~Mhaske, H.~Kedia, S.~Doddapaneni, M.~M. Khapra, P.~Kumar, R.~M. V, and A.~Kunchukuttan, ``Naamapadam: A large-scale named entity annotated data for indic languages,'' 2023. [Online]. Available: \url{https://arxiv.org/abs/2212.10168}
\BIBentrySTDinterwordspacing

\bibitem{Vavekanand2025SindhiNLP}
\BIBentryALTinterwordspacing
S.~Kumar and R.~Vavekanand, ``Multiclass text classifications of sindhi newspaper articles,'' \emph{Preprints}, 2025. [Online]. Available: \url{https://www.preprints.org/frontend/manuscript/d40099f1eed56b67c6f65d138e209557/download_pub}
\BIBentrySTDinterwordspacing

\end{thebibliography}

\end{document}